\pdfoutput=1

\documentclass[11pt]{article}

\usepackage[final]{acl}

\usepackage{times}
\usepackage{latexsym}

\usepackage[T1]{fontenc}

\usepackage[utf8]{inputenc}

\usepackage{microtype}

\usepackage{inconsolata}

\usepackage{graphicx}
\usepackage{url}
\title{SwissGPC v1.0 - The Swiss German Podcasts Corpus}

\author{Samuel Stucki, Mark Cieliebak, Jan Deriu \\
  Centre for Artificial Intelligence \\
  Zurich University of Applied Sciences (ZHAW), Winterthur \\
  \texttt{stku@zhaw.ch, ciel@zhaw.ch, deri@zhaw.ch} \\}

\begin{document}
\maketitle
\begin{abstract}
We present SwissGPC v1.0, the first mid-to-large-scale corpus of spontaneous Swiss German speech, developed to support research in ASR, TTS, dialect identification, and related fields. The dataset consists of links to talk shows and podcasts hosted on Schweizer Radio und Fernsehen and YouTube, which contain approximately 5400 hours of raw audio. After segmentation and weak annotation, nearly 5000 hours of speech were retained, covering the seven major Swiss German dialect regions alongside Standard German. 

We describe the corpus construction methodology, including an automated annotation pipeline, and provide statistics on dialect distribution, token counts, and segmentation characteristics. Unlike existing Swiss German speech corpora, which primarily feature controlled speech, this corpus captures natural, spontaneous conversations, making it a valuable resource for real-world speech applications.

\end{abstract}

\section{Introduction}
Swiss German is a family of dialects spoken in Switzerland and belongs to the Alemannic group of German dialects. It differs from Standard German in phonetics, grammar, vocabulary, and syntax. The dialects vary significantly across regions and are collectively spoken by approximately five million people. Unlike many other dialect groups, Swiss German is widely used in both professional and private settings and additionally serves as an expression and representation of a distinct Swiss nationality in the German-speaking part of the country. While it is primarily a spoken language, the rise of informal digital communication has led to an increase in written Swiss German. However, the absence of standardized orthography and its classification as a low-resource language make data collection for Automatic Speech Recognition (ASR) and other speech processing tasks particularly challenging.

There has been growing interest in the past years in researching ASR tasks on Swiss German dialects, which lead to the creation of several corpora such as the Swiss Parliament Corpus \citep{swiss_parliament_corpus}, SwissDial \citep{swiss_dial_schoenenberger}, SDS-200 \citep{pluss2022sds200}, and STT4SG-350 \cite{pluss2023stt4sg}. These corpora contain between 28 and 343 hours of audio and have since enabled various research endeavours \citep{sicard-etal-2023-spaiche, paonessa-etal-2023-dialect, bollinger2023texttospeechpipelineswissgerman, dolev-etal-2024-whisper}. 

However, these corpora are insufficient for data-intensive tasks such as Text-to-Speech (TTS). This paper thus presents the first version of the "Swiss German Podcasts Corpus (SwissGPC v1.0)", the first mid-to-large-scale\footnote{The dataset can be considered large-scale in the context of Swiss German corpora. However, compared to other languages such as English, German, or Mandarin it is still a small- to medium-sized corpus} corpus for Swiss German: It contains links\footnote{Due to copyright reasons, we can not provide the audio files, but only the links to the websites and the processing pipeline.} to talk shows and podcasts collected from Schweizer Radio und Fernsehen (SRF) and YouTube (YT). These collected data contain approximately 5400 hours of raw audio, including speech from all dialect regions and Standard German. We utilized the 7 dialect regions outlined in \citep{pluss2023stt4sg} to simplify the dialect classification. Only the source links of the utilized shows are released, as we do not possess the legal rights to distribute the audio or the annotated data of both SRF and YT.

\section{Corpus Requirements}
Our primary motivation for creating SwissGPC was to train a Zero-Shot Voice Adaptation Text-to-Speech (TTS) system for Swiss German dialects, for which large amounts of high-quality data are required. The dataset was thus created with the following goals in mind:
\begin{enumerate}
\item The corpus should be sufficiently large with a goal of 4000-5000 hours of primarily Swiss German speech.
\item The corpus must be sufficiently diverse in speakers to provide useful training data for TTS\footnote{Note that this will also be very helpful for downstream ASR tasks.}.
\item The speech must be recorded with a high-quality recording setup.
\item The corpus should cover a diverse set of topics. 
\end{enumerate}

Based on these goals, we decided to collect a large number of dialogues from podcasts that are primarily in Swiss German and to preprocess them to make them applicable for TTS and other speech processing tasks. 

\section{Data Annotation Pipeline}
As outlined in the introduction, we do not have the rights to distribute the audio. We will only publish the links to the podcast sources that comprise the corpus. For SRF podcasts there exists an official API\footnote{https://developer.srgssr.ch/}, while for YouTube, a third-party tool can be used such as pytube to download the files (specifically the pytubefix fork \citep{pytubefix_github}, as the original library is not maintained anymore). Table \ref{tab:srf_corpus_sources} and \ref{tab:yt_corpus_sources} list the podcasts and their online source for SRF and YouTube, respectively. All sources combined link, at the time of publication, to 5404 hours of audio.

\begin{table}[ht]
    \centering
    \small
    \resizebox{0.48\textwidth}{!}{%
    \begin{tabular}{p{4.5cm}|r}
        \hline
        \textbf{SRF Podcast Name} & \textbf{Length (h)}\\ \hline
        \href{https://www.srf.ch/audio/srfglobal}{\#SRFglobal} & 36.97\\ \hline
        \href{https://www.srf.ch/audio/100-sekunden-wissen}{100 Sekunden Wissen} & 186.75 \\ \hline
        \href{https://www.srf.ch/audio/debriefing-404}{Debriefing 404} & 245.14\\ \hline
        \href{https://www.srf.ch/audio/digital-podcast}{Digital Podcast} & 428.05\\ \hline
        \href{https://www.srf.ch/audio/dini-mundart}{Dini Mundart} & 39.39\\ \hline
        \href{https://www.srf.ch/audio/gast-am-mittag}{Gast am Mittag} & 33.14\\ \hline
        \href{https://www.srf.ch/audio/geek-sofa}{Geek-Sofa} & 317.28 \\ \hline
        \href{https://www.srf.ch/audio/srf-wissen}{SRF-Wissen} & 45.05 \\ \hline
        \href{https://www.srf.ch/audio/kultur-talk}{Kultur-Talk} & 55.84\\ \hline
        \href{https://www.srf.ch/audio/literaturclub-zwei-mit-buch}{Literaturclub - Zwei mit Buch} &  31.79\\ \hline
        \href{https://www.srf.ch/audio/medientalk}{Medientalk} & 66.46\\ \hline
        \href{https://www.srf.ch/audio/pipifax/eigene-beduerfnisse-wie-nehme-ich-mir-zeit-fuer-mich-selber-1-20?uuid=8c53e199-78dd-4ae9-8337-f6bc08286967}{Pipifax} & 9.08 \\ \hline
        \href{https://www.srf.ch/audio/podcast-am-pistenrand}{Podcast am Pistenrand} & 18.29  \\ \hline
        \href{https://www.srf.ch/audio/samstagsrundschau}{Samstagsrundschau} & 404.14 \\ \hline
        \href{https://www.srf.ch/audio/sternstunde-philosophie}{Sternstunde Philosophie} & 159.39 \\ \hline
        \href{https://www.srf.ch/audio/sternstunde-religion}{Sternstunde Religion} & 60.82 \\ \hline
        \href{https://www.srf.ch/audio/sykora-gisler}{Sykora Gisler} & 152.22 \\ \hline
        \href{https://www.srf.ch/audio/tagesgespraech}{Tagesgespräch} & 1661.33\\ \hline
        \href{https://www.srf.ch/audio/ufwaermrundi}{Ufwärmrundi} & 60.98 \\ \hline
        \href{https://www.srf.ch/audio/vetters-toene}{Vetters Töne} & 25.42\\ \hline
        \href{https://www.srf.ch/audio/wetterfrage}{Wetterfrage} & 67.68 \\ \hline
        \href{https://www.srf.ch/audio/wirtschaftswoche}{Wirtschaftswoche} & 122.30 \\ \hline
        \href{https://www.srf.ch/audio/wissenschaftsmagazin}{Wissenschaftsmagazin} & 393.61 \\ \hline
        \href{https://www.srf.ch/audio/zivadiliring}{Zivadiliring} & 50.03\\ \hline
        \href{https://www.srf.ch/audio/zytlupe}{Zytlupe} & 44.74\\ \hline\hline
        \textbf{Total} & 4715.87 \\ \hline

    \end{tabular}}
    \caption{List of SRF podcasts, links to the source, and hours of raw audio.}
    \label{tab:srf_corpus_sources}
\end{table}
        
\begin{table}[ht]
    \centering
    \small
    \resizebox{0.48\textwidth}{!}{%
    \begin{tabular}{p{4.5cm}|r}
        \hline
        \textbf{YouTube Podcast Name} & \textbf{Length (h)}\\ \hline
        \href{https://www.youtube.com/playlist?list=PLAD8a6PKLsRhHc-uS6fA6HTDijwE5Uwju}{Auf Bewährung - Leben mit Gefängnis} & 3.00 \\ \hline
        \href{https://www.youtube.com/playlist?list=PLyWje_91744G6UAsfHjTLWDtejJdHmuYv}{Berner Jugendtreff} & 127.80 \\ \hline
        \href{https://www.youtube.com/playlist?list=PLCospSPttrrVSk0N5Mqj1dveKZtDZNOAl}{Ein Buch Ein Tee} & 3.73 \\ \hline
        \href{https://www.youtube.com/playlist?list=PL5ZbqYujTUkVmNCGMP4e0yFVhY8P5EC73}{expectations - geplant und ungeplant kinderfrei} & 16.84 \\ \hline
        \href{https://www.youtube.com/playlist?list=PL356t1Y2d_AXycvLzBF1n8ee0uM4pw9JX}{Fadegrad} & 49.95\\ \hline
        \href{https://www.youtube.com/playlist?list=PLf-k85Nq3_j-glR2im1SZv_BxqzdYdENk}{Feel Good Podcast} & 319.60  \\ \hline
        \href{https://www.youtube.com/playlist?list=PLGJjtm2tSyhQXU-_N2YkfqCffXhY6UHNe}{Finanz Fabio} & 58.44 \\ \hline
        \href{https://www.youtube.com/playlist?list=PLKaFe_fDMhQNbWvnJGC6HArb285ZUdGbz}{Scho ghört} & 23.45 \\ \hline
        \href{https://www.youtube.com/playlist?list=PL3D2QP2F5r9VDSj6YQb6Ihr_63Gxtm4L5}{Sexologie - Wissen macht Lust} & 15.41 \\ \hline
        \href{https://www.youtube.com/playlist?list=PLPtjJ0sjI3yzhNtZUBY0_e462_gKtr90V}{Über den Bücherrand} & 14.53  \\ \hline
        \href{https://www.youtube.com/playlist?list=PLM4IdPP-Tx3W84w1GB8cn33GnuIGcqaeP}{Ungerwegs Daheim} & 38.67 \\ \hline
        \href{https://www.youtube.com/playlist?list=PLtTxFB6b5Pljl4RU6vimwfQpV490K6SQe}{Wir müssen reden - Public Eye spricht Klartext} & 17.52 \\ \hline \hline
        \textbf{Total} & 688.93 \\ \hline
    \end{tabular}}
    \caption{List of YouTube podcasts, links to the source, and hours of raw audio.}
    \label{tab:yt_corpus_sources}
\end{table}

\begin{figure}[t]
  \includegraphics[width=\columnwidth]{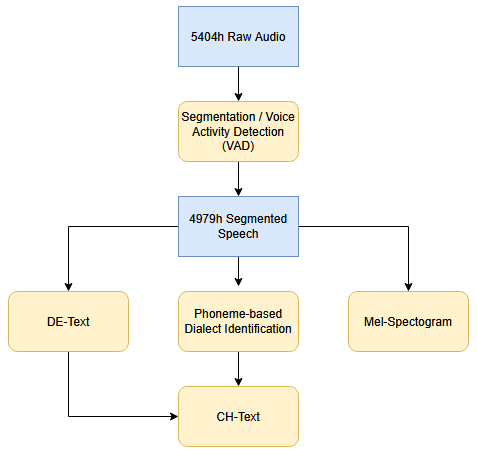}
  \caption{Automated Data Annotation Pipeline}
  \label{fig:data_annotation_pipeline}
\end{figure}

The data was weakly annotated using an automated pipeline, visualized in Figure \ref{fig:data_annotation_pipeline}. First, the raw audio was diarized and segmented on a speaker basis using pyannote \citep{Bredin23}. The diarization step only tags actual speech, leading to silent and music segments being implicitly removed. The samples, containing only a single speaker based on the diarization, were cut to be between 2 and 15 seconds long. The time range was chosen to allow diverse sampling of shorter and longer segments, and additionally due to models downstream that required different lengths of audio for transcription or training. This resulted in a reduction of 7.84\% from 5404 hours of raw audio to 4979 hours of actual speech with 1.76M unique samples. The segmented audio was then transcribed to Standard German since it has a standardized orthography. The transcription was performed using whisper-v3 \citep{radford2022robustspeechrecognitionlargescale} for its high performance in translating Swiss German speech to Standard German text \citep{paonessa-etal-2024-whisper}. Using the approach of \citep{bolliger2024automatische}, we applied a wav2vec2 phoneme transcriber \citep{NEURIPS2020_92d1e1eb,xu22b_interspeech}. We classified the generated phoneme sequences with a Naïve Bayes n-gram classifier trained on the phonemicized STT4SG-350 corpus \citep{pluss2023stt4sg} and an additional Standard German CommonVoice \citep{ardila2020commonvoice} subset for Dialect Identification (DID). In total 8 different regions were thus used in classification: Basel, Bern, Central CH, Eastern CH, Grisons, Valais, Zurich, and Standard German.
Further enrichment processes included the generation of Swiss German text using \citep{bollinger2023texttospeechpipelineswissgerman} and the creation of Mel-Spectrogram of the audio samples using \citep{mcfee_2024_11192913}.

\subsection{Pipeline Evaluation}
In order to ensure high quality of the automated annotations, we performed an evaluation of the performance of the individual steps of the annotation pipeline. This section presents the evaluation results. The \textit{Zivadiliring} podcast\footnote{https://www.srf.ch/audio/zivadiliring} was selected for all evaluations due to its moderate size (approximately 50 hours of raw audio), exclusive use of Swiss German, minimal guest appearances, and the known dialects of its hosts—one from Eastern Switzerland and two from Zurich. These characteristics make it a representative sample of other podcasts.

The diarization was evaluated on a single episode lasting 42 minutes and 38 seconds, using \citep{elan2024, brugman-russel-2004-annotating} for manual annotation. The diarization pipeline achieved a Diarization Error Rate (DER) of 14.1\%, which is comparable to its performance on the AISHELL-4 corpus \cite{aishell_2017}, where it reached 12.2\%. The Standard German transcription was evaluated by manually transcribing 100 randomly selected audio samples, achieving a Word Error Rate (WER) of 0.30\(\pm\)0.264. Example transcriptions are provided in Table \ref{tab:de_text_eval}. The observed deletions and substitutions can be attributed to the numerous linguistic differences between Swiss German and Standard German. These include the omission of past tense forms, instead preferring the perfect tense, as well as variations in auxiliary verbs, grammatical structures, and the use of Helvetisms or loanwords that either do not exist in Standard German or carry different meanings. Lastly, there is the inherent loss of information when transcribing the audio automatically from Swiss German to Standard German using whisper.

\begin{table}[ht]
    \centering
    \small
    \resizebox{0.48\textwidth}{!}{%
    \begin{tabular}{|p{1cm}|p{2.5cm}|p{2.5cm}|}
    \hline
    \textbf{Dialect} & \textbf{Hypothesis} & \textbf{Reference} \\ \hline
    Eastern CH  &  Und dann ist quasi die Idee, wenn du als Burning Man gehst, dass du etwas wie einen Provider machst. & Dann ist quasi die Idee auch, dass du, wenn du an das Burning Man gehst, etwas providest. \\ \hline
    Zurich                 &  Aber es ist nicht gescheitert. Nein, ich bin ja so hyper-emotional. Dann verplatzt es mich und dann bin ich aber wieder ruhig nach vier Sekunden. Aber ich habe dann schon wahrscheinlich ein bisschen umgewettert.        & Aber es ist nicht gescheitert an der Wäsche. Nein, ich bin ja schon, ich bin ja so Hyperemotional, dann verplatzt es mich, dann bin ich aber auch wieder ruhig nach vier Sekunden. Aber habe dann wahrscheinlich schon herumgeflucht.            \\ \hline
    Zurich                  &  Oder was ist er? Weisst du, mit dem Rettchen wüsstest du, über was wir reden. Was ist er gestern gewesen? Was ist er heute?  & Oder was ist er? Weisst du damit wir wissen über was wir reden. Was war er gestern? Was ist er heute?                 \\ \hline
    \end{tabular}}
    \caption{Comparison of generated and manual annotated Standard German sentences}
    \label{tab:de_text_eval}
\end{table}

The Naïve Bayes classifier from \citep{bolliger2024automatische} was retrained with an additional class for Standard German using the CommonVoice corpus \citep{ardila2020commonvoice}. A total of 30 hours of audio was sampled from CommonVoice, ensuring an age and gender distribution similar to that of the phonemicized STT4SG-350 corpus \citep{pluss2023stt4sg}, where each dialect region consists of 30 hours of speech. The classifier achieved a macro F1-score of 0.88 across the eight regions. When applied to the \textit{Zivadiliring} episodes, nearly two-thirds of all samples were classified as Zurich and one-third as Eastern Switzerland, aligning with the hosts' origins. Additionally, in an episode where one of the hosts was replaced by a guest from Basel, the classifier correctly identified the samples as Basel.

Lastly, the Swiss German transcription of the same 100 samples used in the Standard German evaluation resulted in a Word Error Rate (WER) of 0.639\(\pm\)0.253. This high error rate is primarily attributed to the lack of a standardized writing system for Swiss German. Example transcriptions are provided in Table \ref{tab:ch_text_eval}.

\begin{table}[ht]
    \centering
    \small
    \resizebox{0.48\textwidth}{!}{%
    \begin{tabular}{|p{1cm}|p{2.5cm}|p{2.5cm}|}
    \hline
    \textbf{Dialect} & \textbf{Hypothesis} & \textbf{Reference} \\ \hline
    Zurich & Si kännt scho mal din Name, fast. Er isch Content Creator, er isch berüehmt im Internet und er isch super. & Sie kennt scho mal din Name, fast. Er isch Content-Creator, er isch berüehmt im Internet und er isch \\ \hline
    Eastern CH & I mein, wa de Onur alles seit. Nur will me zemme wohned isch jetzt nöd de Informationsfluss. & Ich meine, was dä Onur alles seit. Nur will mir zemme wohnet isch ezt do nöd de Informationsfluss. \\ \hline
    Zurich & Drum händs so gfunde, ja du bisch irgendwie d'Muetter und denn au irgendwie nöd. Ich glaub, es git nöd die definiert Rolle. Aber ich han so gfunde d'Klaschtante isch no härzig. & Drum hät si d Mueter gfunde, dass si das au nöd gseh hät, dass Klatschstunde no härzig isch. \\ \hline
    
    \end{tabular}}
    \caption{Comparison of generated and manual annotated Swiss German sentences}
    \label{tab:ch_text_eval}
\end{table}

\section{Corpus Statistics}
\begin{figure}[ht]
  \includegraphics[width=\columnwidth]{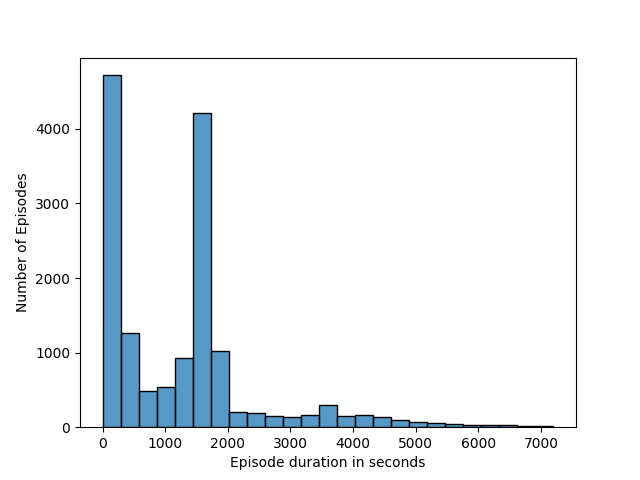}
  \caption{Distribution of episode durations in seconds of all podcasts in the corpus. Outliers (\(n=32\)) with length > 7200 seconds are not shown.}
  \label{fig:distro_ep_lengths}
\end{figure}

\emph{Raw Data.} The raw audio is sourced from 25 SRF and 12 YouTube podcasts, comprising 15171 individual episodes with an average length of 1277.28 seconds (21.28 minutes). Episode durations are unevenly distributed, visualized in Figure \ref{fig:distro_ep_lengths}, forming two distinct peaks: one between 100 and 200 seconds and another between 1,600 and 1,800 seconds. The first peak is primarily due to the podcast \textit{100 Sekunden Wissen}, in which hosts provide information about various topics in around 100 seconds. In general, most podcasts produce episodes ranging from 20 to 30 minutes in length, as seen in Figure \ref{fig:ep_vs_avg_dur_eps}. 

The largest podcast is \textit{Tagesgespräch} with 1661.33 hours of raw audio, comprising nearly 31\% of the total dataset, clearly visible in Figure \ref{fig:ep_vs_total_dur}. On average there are 410 episodes in a podcast, while the median is significantly lower at 104. Outlier episodes (\(>7200s, n=32\)) were typically special episodes, such as yearly recaps, video game playthroughs, or guest interviews. The longest episode in the dataset lasted 13846 seconds (3 hours and 50 minutes), while the shortest was just 19 seconds.

\emph{Filtering.} The filtering step, where we remove, for instance, samples with music only, reduced the data from 5404 hours of audio to 4979 hours of speech, which was segmented into 1.76M samples 

\begin{figure}[ht]
  \includegraphics[width=\columnwidth]{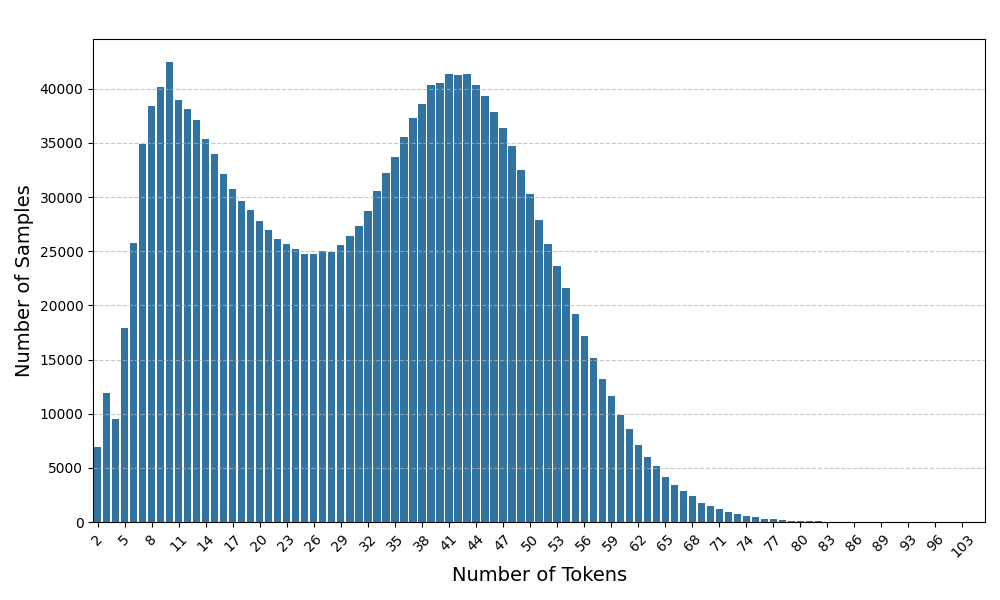}
  \caption{Standard German token distribution of segmented audio samples.}
  \label{fig:standard_german_token_distro}
\end{figure}

\emph{Token Counts.} After filtering, the data contains 55.85M tokens, calculated using spaCy \citep{Honnibal_spaCy_Industrial-strength_Natural_2020}. The token distribution is shown in \ref{fig:standard_german_token_distro}. Since we did not segment the data on the sentence level, this led to a bimodal distribution of the tokens, visualized in Figure \ref{fig:standard_german_token_distro}, with a large concentration of samples at 15 seconds. Training of a TTS model downstream then led to longer segments being generated better than shorter ones. Future work may improve this. The first peak with token counts of between 7 and 14 could be explained by the characteristics of spontaneous speech in a podcast setting, in which hosts often interrupt each other in turns or simultaneously, leading to short segments of speech from individual hosts. The second peak with token counts between 40 and 53 can be explained by the generally more information-dense segments in podcasts or shows, where hosts have a monologue telling a story, reading a book or letter, or similar. Additionally, it was found that very short (< 7 tokens) and very large (\(\geq\) 65 tokens) samples were often erroneous or incoherent translations by whisper, either due to complex audio or simple mistranslations.

\emph{Dialects.} At the dialect level, the corpus is highly unbalanced: the two largest regions, Standard German and Zurich, account for 57.53\% of audio in the dataset, while the smallest region, Valais, represents only 0.79\%. Table \ref{tab:srf_statistics} provides further insight into the dialect distributions. Additionally, it was observed that Standard German tended to have segments with larger token counts than other dialect regions. This can be attributed to SRF broadcasting more formal and information-dense segments such as science, philosophy, and news programs in Standard German rather than Swiss German. This ensures that all Swiss residents can understand the content regardless of their familiarity with Swiss German dialects. The moderators of these programs are not required to be from Switzerland and could thus originate from Germany, Austria, or another German-speaking region. We hypothesized that the pronunciation of Swiss German speakers using Standard German (i.e., Swiss Standard German) may have a beneficial effect on model training. However, as we are currently unable to distinguish between them, both are grouped under the Standard German label and kept in the dataset.

\begin{table}[t!]
\centering
\small
\resizebox{0.48\textwidth}{!}{%
\begin{tabular}{l|rrrr}
\hline
\textbf{Region}   & \textbf{Samples (K)} & \textbf{Length (h)} & \textbf{\% of Dataset} & \textbf{Tokens (M)} \\ \hline
Basel               & 179   & 460.81    & 9.25\%    & 5.35  \\ 
Bern                & 293	& 771.38    & 15.49\%   & 8.98  \\ 
German              & 538   & 1685.72   & 33.86\%   & 17.23  \\
Grisons             & 57	& 151.33    & 3.04\%   & 1.74   \\ 
Central CH          & 121	& 341.22    & 6.85\%   & 3.95    \\ 
Eastern CH          & 121	& 350.60    & 7.04\%   & 4.00    \\ 
Valais              & 15	& 39.46     & 0.79\%   & 0.43    \\ 
Zurich              & 440	& 1178.58   & 23.67\%  & 14.13   \\ \hline
Total               & 1767	& 4979.09   & 100.00\%    & 55.81  \\ \hline 

\end{tabular}}
\caption{Corpus statistics by dialect concerning number of samples, duration, percentage of total duration, and number of tokens.}
\label{tab:srf_statistics}
\end{table}

\section{Potential Use Cases}
The Swiss German Podcasts Corpus can be a valuable resource for various NLP tasks, particularly for Swiss German. Unlike many existing datasets that focus on scripted or carefully controlled speech, our corpus contains spontaneous, natural, and uncontrolled speech. This makes it particularly useful for real-world applications where speech is often erratic, featuring hesitations, interjections, interruptions, and overlapping speakers. The large size of the corpus and the weak annotation make it particularly useful for weakly supervised learning approaches. An example task where this approach yielded very good results is in \emph{Voice Adaptation for Swiss German dialects} using the XTTSv2 architecture \citep{casanova24_interspeech}. Since the corpus contains a mix of Swiss German and Standard German, it can also serve as an excellent resource for training \emph{Swiss German–to–Standard German machine translation models}. Such models can bridge the gap between spoken dialects and formal written language, enabling better transcription and translation.

\section{Corpus Access}
SwissGPC v1.0 is accessible via \href{https://github.com/stucksam/SwissGPC}{GitHub}. The corpus will include:
\begin{enumerate}
\item A comprehensive list of links to all podcasts sourced from SRF and YouTube
\item The code for both downloading any podcast from SRF and YT and the automated annotation pipeline
\end{enumerate}

\section{Conclusion}
We have presented the Swiss German Podcast Corpus (SwissGPC v1.0), the first mid-to-large-scale Swiss German speech corpus comprising approximately 5400 hours of raw audio (4979 hours of speech after data cleaning). While the audio can not be released due to licensing concerns, we have provided references to individual podcasts, including an approach for downloading the audio. Additionally, we defined an automated annotation pipeline to weakly label the data for downstream use. 

We are convinced that SwissGPC will enable interesting research in the Swiss German speech processing space, and we are excited to see applications utilizing it.

\section*{Limitations}
The corpus represents a snapshot in time of the selected podcasts. Shows may release new episodes, remove existing ones, change name or location, be discontinued, or be taken offline as a whole. As a result, reproducing the results given here may prove challenging.

SwissGPC v1.0 is highly imbalanced on a dialectal basis, and future work may seek more audio from under-represented regions and add it to the corpus. 

The list of podcasts from SRF is not exhaustive, as during the writing of this paper additional podcasts were found that could be utilized. Additionally, it should also be possible to crawl TV shows from SRF, such as \textit{SRF bide lüt}, \textit{Arena}, and more via their website\footnote{https://www.srf.ch/play/tv/sendungen} or YouTube channel\footnote{https://www.youtube.com/@srfdoku}, increasing the size of the corpus further. 


\bibliography{custom}

\appendix

\section{Corpus Statistics}\label{sec:appendix_stats}

\begin{figure}[ht]
  \includegraphics[width=\columnwidth]{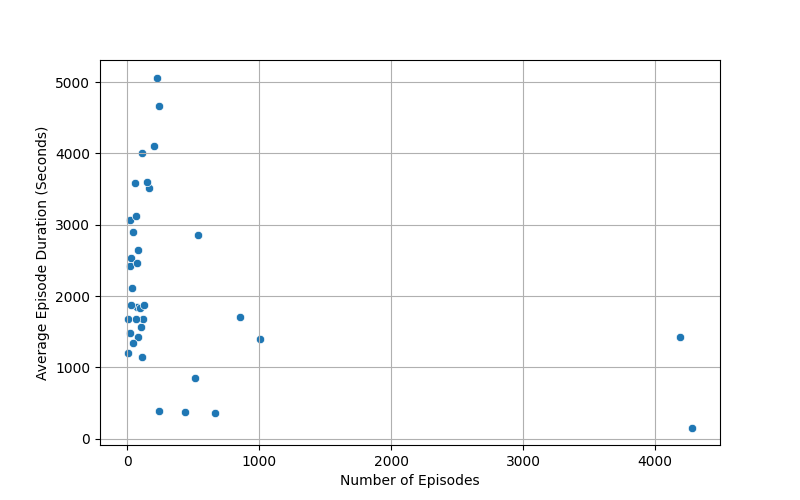}
  \caption{Comparison of total number of episodes in a podcast to the average duration per episode in the podcast.}
  \label{fig:ep_vs_avg_dur_eps}
\end{figure}

\begin{figure}[ht]
  \includegraphics[width=\columnwidth]{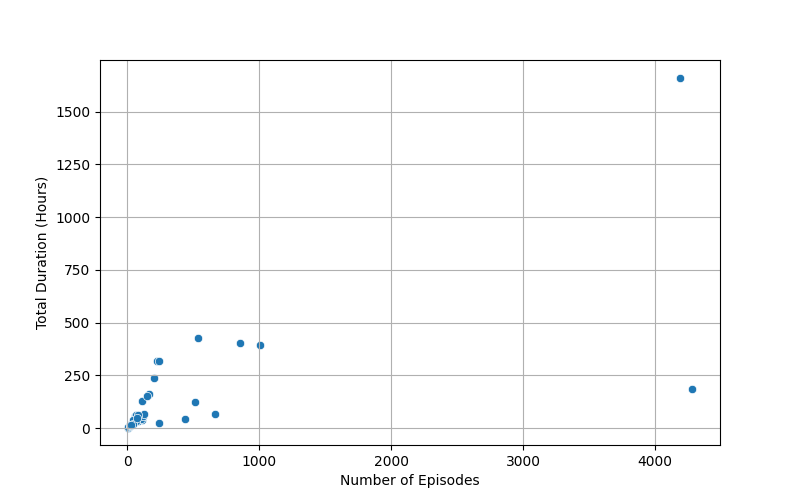}
  \caption{Comparison of the total number of episodes in a podcast to the total duration of all episodes combined in the podcast.}
  \label{fig:ep_vs_total_dur}
\end{figure}

\begin{table}[ht]
    \centering
    \small
    \resizebox{0.48\textwidth}{!}{%
    \begin{tabular}{|p{2.5cm}|p{2.5cm}|p{1cm}|p{1.5cm}|}
    \hline
    \textbf{DE-Text} & \textbf{CH-Text} & \textbf{Dialect} & \textbf{Podcast} \\ \hline
    Die Kosten steigen natürlich dieses Jahr zwischen 6 und 7 Prozent, je nach Leistungserbringerbereich. Aber letztes Jahr hatte man schon Defizite. Also schon letztes Jahr haben die Prämieneinnahmen die Ausgaben nicht gedeckt. Dieses Jahr wird es noch schlimmer sein. Das nächste & D Kösta stigen natürlich das Johr zwüscha sechs und sieba Prozent, je noch Leistigserbringerberich. Aber letschts Johr hend d Prämieneinahma d Usgaba nit bedeckt kah. & Grisons & Samstags-rundschau \\ \hline
    grossen Deutschschweizer Massenmedien, die noch eine regelmässige Gamekritik gemacht haben. Alle anderen haben das schon viel länger aufgegeben als wir. Und auch bei den Spezialisten, die sich jetzt spezifisch für Gamer & Grosse Dütschschwizer Massemedia, wo no e regelmässigi Gameskritik gmacht händ. Alli andere händ das scho vill länger ufgeh wie mir. Und au bi de Spezialiste, wo sich jetzt spezifisch für Games. & Zurich & Geek-Sofa \\ \hline
    Es war eine Erleichterung, nachdem die UBS angekündigt hat, dass sie auf die staatlichen Garantien verzichtet, die wir im März sprechen mussten. Ohne viel Enthusiasmus. & Es isch e Erliechterig gsi, nochdem dUBS aakündigt het, dass sie uf di staatliche Garantie vozichtet hend, wome im März sproche müend. & Eastern CH & Tages-gespräch \\ \hline               
    nur noch mit Katalysatoren zulassen, so würde man längerfristig den Schadstoffausstoss massiv beschränken können. & Nor no met Katalysatore zueloh, so wörd mer längerfrestigi de Schadstoffusstoss massiv chönne beschränke. & Central CH & 100 Sekunden Wissen \\ \hline

    \end{tabular}}
    \caption{Examples of segmented samples in the corpus.}
    \label{tab:example_samples_podcasts}
\end{table}

\end{document}